\documentclass[11pt,a4paper]{article}
\usepackage[hyperref]{emnlp-ijcnlp-2019}
\usepackage{amsmath}
\usepackage{amssymb}
\usepackage{amsfonts}
\usepackage{array}
\usepackage{bm}
\usepackage{booktabs}
\usepackage[labelfont=bf,skip=5pt]{caption}
\usepackage{cases}
\usepackage{color}
\usepackage{diagbox}
\usepackage{enumitem}
\usepackage{graphicx}
\usepackage{latexsym}
\usepackage{makecell}
\usepackage[mathscr]{euscript}
\usepackage{mathtools}
\usepackage{multirow}
\usepackage{paralist}
\usepackage{sidecap}
\usepackage{soul}
\usepackage{subcaption}
\usepackage{tabu}
\usepackage{tabulary}
\usepackage{times}
\usepackage{url}
\usepackage{verbatim}
\usepackage{xspace}

\newcommand{\sect}[1]{\S\ref{#1}}
\newcommand{\seq}[1]{\boldsymbol{#1}}

\newcommand{\ai}{\texttt{AllenNLP}}
\newcommand{\cwrs}{\textsc{cwr}s\xspace}
\newcommand{\cwr}{\textsc{cwr}\xspace}
\newcommand{\msync}{\textbf{mSynC}\xspace}
\newcommand{\elmot}{ELMo-transformer\xspace}

\aclfinalcopy % Uncomment this line for the final submission

\newcommand{\ensuretext}[1]{#1}
\newcommand{\marker}[2]{\ensuremath{^{\textsc{#1}}_{\textsc{#2}}}}
\newcommand{\draftonly}[1]{#1}
\renewcommand{\draftonly}[1]{}
\newcommand{\draftcomment}[3]{\draftonly{\ensuretext{\textcolor{#3}{[#1 #2]}}}}
\newcommand{\swabha}[1]{\draftcomment{\marker{S}{S}}{#1}{purple}}

\newcommand{\matt}[1]{\draftcomment{\marker{M}{P}}{#1}{brown}}

\newcommand{\todo}[1]{\draftcomment{TODO}{#1}{red}}

\title{Shallow Syntax in Deep Water}

\author{ 
    Swabha Swayamdipta$^\spadesuit$\thanks{\quad Work done during an internship at the Allen Institute for Artificial Intelligence.} \quad
	Matthew Peters$^\clubsuit$  \\ 
	\bf Brendan Roof$^\clubsuit$ \quad
	Chris Dyer$^\heartsuit$ \quad
	Noah A. Smith$^{\diamondsuit \clubsuit}$ \\\\
	$^\spadesuit$Language Technologies Institute, Carnegie Mellon University \\
	$^\clubsuit$Allen Institute for Artificial Intelligence \\
	$^\diamondsuit$Paul G. Allen School of Computer Science \& Engineering, University of Washington \\
	    $^\heartsuit$Google DeepMind \\
{\tt \{swabhas,matthewp,brendanr\}@allenai.org}
	}

\date{}

\begin{document}
\maketitle

\begin{abstract}
Shallow syntax provides an approximation of phrase-syntactic structure of sentences; it can be produced with high accuracy, and is computationally cheap to obtain.
We investigate the role of shallow syntax-aware representations for NLP tasks using two techniques.
First, we enhance the ELMo architecture \citep{Peters:18b} to allow pretraining on predicted shallow syntactic parses, instead of just raw text, so that contextual embeddings make use of shallow syntactic context.
Our second method involves shallow syntactic features obtained automatically on downstream task data.
Neither approach leads to a significant gain on any of the four downstream tasks we considered relative to ELMo-only baselines.
Further analysis using black-box probes from \citet{Liu:19} confirms that our shallow-syntax-aware contextual embeddings do not transfer to linguistic tasks any more easily than ELMo's embeddings.
We take these findings as evidence that ELMo-style pretraining discovers representations which make additional awareness of shallow syntax redundant.
\end{abstract}

\section{Introduction}
\label{sec:intro}

\begin{figure}
\footnotesize
\centering
    \includegraphics[scale=0.37]{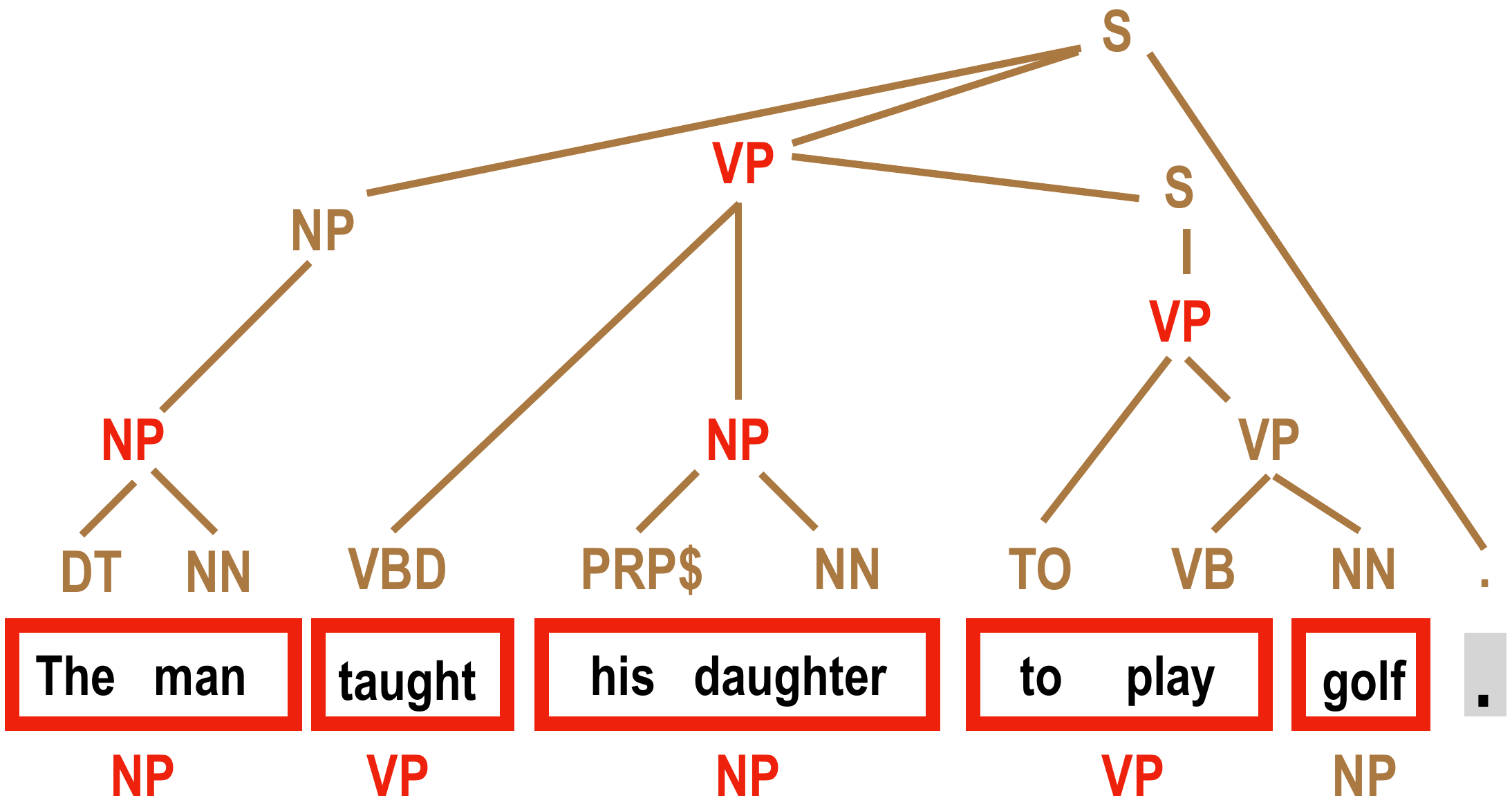}
    \caption{
    A sentence with its phrase-syntactic tree (brown) and shallow syntactic (chunk) annotations (red). 
    Nodes in the tree which percolate down as chunk labels are in red.
    Not all tokens in the sentence get chunk labels; e.g., punctuation is not part of a chunk.}
    \label{fig:chunk_tree}
\end{figure}

The NLP community is revisiting the role of linguistic structure in applications with the advent of contextual word representations (\cwrs) derived from pretraining language models on large corpora~\cite{Peters:18,Radford:18,Howard:18,Devlin:18}.
Recent work has shown that downstream task performance may benefit from explicitly injecting a syntactic inductive bias into model architectures \citep{Kuncoro:18}, even when \cwrs are also used \cite{Strubell:18}.
However, high quality linguistic structure annotation at a large scale remains expensive---a trade-off needs to be made between the quality of the annotations and the computational expense of obtaining them.
Shallow syntactic structures (\citealp{Abney:91}; also called chunk sequences) offer a viable middle ground, by providing a flat, non-hierarchical approximation to phrase-syntactic trees (see Fig.~\ref{fig:chunk_tree} for an example).
These structures can be obtained efficiently, and with high accuracy, using sequence labelers.
In this paper we consider shallow syntax to be a proxy for linguistic structure. 

While shallow syntactic chunks are almost as ubiquitous as part-of-speech tags in standard NLP pipelines \cite{Jurafsky:00}, their relative merits in the presence of \cwrs remain unclear.
We investigate the role of these structures using two methods.
First, we enhance the ELMo architecture \citep{Peters:18b} to allow pretraining on predicted shallow syntactic parses, instead of just raw text, so that contextual embeddings make use of shallow syntactic context (\sect{sec:pretraining}).
Our second method involves classical addition of chunk features to \cwr-infused architectures for four different downstream tasks (\S\ref{sec:shallow_features}).
Shallow syntactic information is obtained automatically using a highly accurate model (97\% $F_1$ on standard benchmarks).
In both settings, we observe only modest gains on three of the four downstream tasks relative to ELMo-only baselines (\sect{sec:experiments}).

Recent work has probed the knowledge encoded in \cwrs and found they capture a surprisingly large amount of syntax~\cite{Blevins:18,Liu:19,Tenney:18}. 
We further examine the contextual embeddings obtained from the enhanced architecture and a shallow syntactic context, using black-box probes from \citet{Liu:19}.
Our analysis indicates that our shallow-syntax-aware contextual embeddings do not transfer to linguistic tasks any more easily than ELMo embeddings (\sect{sec:probes}).

Overall, our findings show that while shallow syntax can be somewhat useful,  ELMo-style pretraining discovers representations which make \textit{additional} awareness of shallow syntax largely redundant.

\section{Pretraining with Shallow Syntactic Annotations} 
\label{sec:pretraining}

We briefly review the shallow syntactic structures used in this work, and then present a model architecture to obtain e\textbf{m}beddings from shallow \textbf{Syn}tactic \textbf{C}ontext (\msync).

\subsection{Shallow Syntax}
\label{sec:chunks}

Base phrase chunking is a cheap sequence-labeling--based alternative to full syntactic parsing, where the sequence consists of non-overlapping labeled segments (Fig.~\ref{fig:chunk_tree} includes an example.)
Full syntactic trees can be converted into such shallow syntactic chunk sequences using a deterministic procedure \cite{Jurafsky:00}.
\citet{Tjong:00} offered a rule-based transformation deriving non-overlapping chunks from phrase-structure trees as found in the Penn Treebank \citep{Marcus:93}.
The procedure percolates some syntactic phrase nodes from a phrase-syntactic tree to the phrase in the leaves of the tree.
All overlapping embedded phrases are then removed, and the remainder of the phrase gets the percolated label---this usually corresponds to the head word of the phrase.

In order to obtain shallow syntactic annotations on a large corpus, we train a BiLSTM-CRF model \cite{Lample:16,Peters:17}, which achieves 97\% $F_1$ on the CoNLL 2000 benchmark test set.
The training data is obtained from the CoNLL 2000 shared task \cite{Tjong:00}, as well as the remaining sections (except \S 23 and \S 20) of the Penn Treebank, using the official script for chunk generation.\footnote{\url{https://www.clips.uantwerpen.be/conll2000/chunking/}} 
The standard task definition from the shared task includes eleven chunk labels, as shown in Table \ref{tab:chunk_labels}.

\begin{table}[tbh]
	\centering
	\begin{tabulary}{\columnwidth}{@{}l   rr@{}}
		\toprule
        Label  & \% Occurrence \\%& Average Width
        \midrule
        
        Noun Phrase (\texttt{NP}) & 51.7\\
		Verb Phrase (\texttt{VP}) & 20.0\\
		Prepositional Phrase (\texttt{PP}) & 19.8 \\
		Adverbial Phrase (\texttt{ADVP}) & 3.7\\
		Subordinate Clause (\texttt{SBAR}) & 2.1\\
		Adjective Phrase (\texttt{ADJP}) & 1.9\\
		\midrule[0.03em]
		
		Verb Particles (\texttt{PRT}) & 0.5\\
		Conjunctive Phrase (\texttt{CONJ}) & 0.06\\
		Interjective Phrase (\texttt{INTJ}) & 0.03\\
		List Marker (\texttt{LST}) & 0.01\\
		Unlike Coordination Phrase (\texttt{UCP}) & 0.002 \\

		\bottomrule
	\end{tabulary}
	\caption{Shallow syntactic chunk phrase types from CoNLL 2000 shared task \citep{Tjong:00} and their occurrence \% in the training data. \swabha{Also add averg width and frequency of occurrence}}
	\label{tab:chunk_labels}
\end{table}

\subsection{Pretraining Objective}
\label{sec:training}

Traditional language models are estimated to maximize the likelihood of each word $x_i$ given the words that precede it, $p(x_i \mid \seq{x}_{<i})$.  
Given a corpus that is annotated with shallow syntax, we propose to condition on both the preceding words \emph{and} their annotations.  

\begin{figure}[tbh]
\footnotesize
\centering
    \includegraphics[scale=0.4]{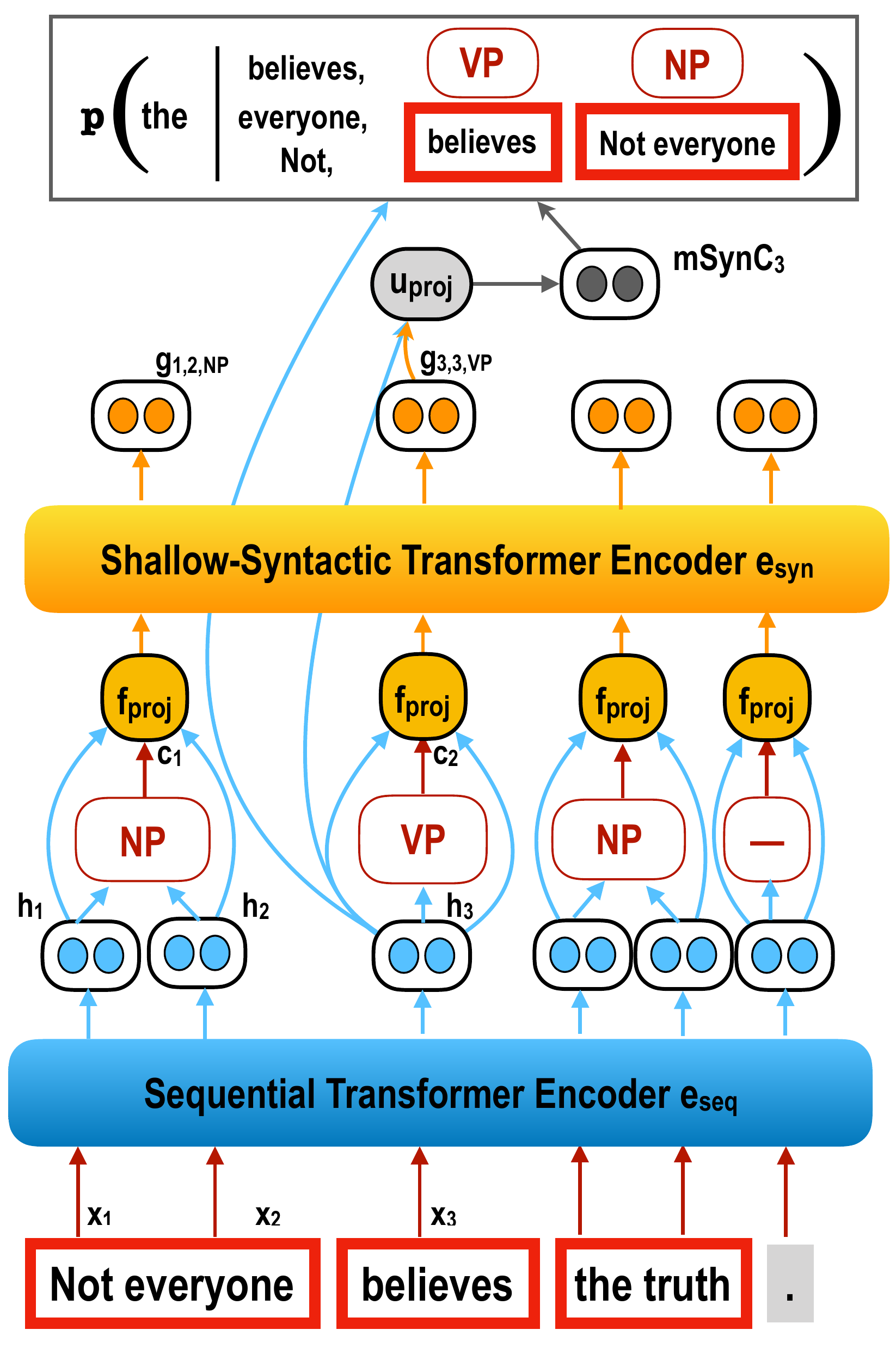}
    \caption{
    Model architecture for pretraining with shallow syntax.
    A sequential encoder converts the raw text into \cwrs(shown in blue).
    Observed shallow syntactic structure (chunk boundaries and labels, shown in red) are combined with these \cwrs in a shallow syntactic encoder to get contextualized representations for chunks (shown in orange).
    Both representations are passed through a projection layer to get \msync embeddings (details shown only in some positions, for clarity), used both for computing the data likelihood, as shown, as well as in downstream tasks.
    }
    \label{fig:chunks}
\end{figure}

We associate with each word $x_i$ three additional variables (denoted $c_i$):  the indices of the beginning and end of the last completed chunk \emph{before} $x_i$, and its label.  
For example, in Fig.~\ref{fig:chunks}, $c_4=\langle 3, 3, \text{VP}\rangle$ for $x_4=\text{the}$.
Chunks, $\seq{c}$ are only used as conditioning context via $p(x_i \mid \seq{x}_{<i}, \seq{c}_{\leqslant i})$;
% \begin{align}
% \label{eq}
%     p(x_i \mid \seq{x}_{<i}, \seq{c}_{\leqslant i});
% \end{align}
they are not predicted.\footnote{A different objective could consider predicting the next chunks, along with the next word. 
However, this chunker would have access to strictly less information than usual, since the entire sentence would no longer be available.}
Because the $\seq{c}$ labels depend on the entire sentence through the CRF chunker, conditioning each word's probability on any $\seq{c}_i$ means that our model is, strictly speaking, not a language model, and it can no longer be meaningfully evaluated using perplexity.

A right-to-left model is constructed analogously, conditioning on  $\seq{c}_{\geqslant i}$ alongside $\seq{x}_{>i}$.  
Following \citet{Peters:18}, we use a joint objective maximizing data likelihood objectives in both directions, with shared softmax parameters. 

\subsection{Pretraining Model Architecture}
\label{sec:architecture}

Our model uses two encoders:  $e_{\mathit{seq}}$ for encoding the sequential history ($\seq{x}_{<i}$), and $e_{\mathit{syn}}$ for shallow syntactic (chunk) history ($\seq{c}_{\leqslant i}$).
For both, we use transformers \cite{Vaswani:17}, which consist of large feedforward networks equipped with multiheaded self-attention mechanisms.
\swabha{{While transformers are shown to have slightly lower performance on downstream tasks than RNNs for pretraining \cwrs \cite{Peters:18b}, this model trains almost twice as fast on two NVIDIA Tesla V100s, and is hence cost-effective.} 
\matt{I'd just remove this footnote as BERT, GPT use transformers}}

As inputs to $e_{\mathit{seq}}$, we use a context-independent embedding, obtained from a CNN character encoder~\cite{Kim:2016:CNL:3016100.3016285} for each token $x_i$.
The outputs $\seq{h}_i$ from $e_{\mathit{seq}}$ represent words in context.

Next, we build representations for (observed) chunks in the sentence by concatenating a learned embedding for the chunk label with $\seq{h}$s for the boundaries and applying a linear projection ($f_\mathit{proj}$).
The output from $f_\mathit{proj}$ is input to $e_{\mathit{syn}}$, the shallow syntactic encoder, and results in contextualized chunk representations, $\seq{g}$.
Note that the number of chunks in the sentence is less than or equal to the number of tokens.

Each $\seq{h}_i$ is now concatentated with $\seq{g}_{c_i}$, where $\seq{g}_{c_i}$ corresponds to $c_i$, the last chunk before position $i$.
Finally, the output is given by $\mbox{\msync}_i = {u}_\mathit{proj}(\seq{h}_i, \seq{g}_{c_i}) = \seq{W}^\top[\seq{h}_i; \seq{g}_{c_i}]$, where $\seq{W}$ is a model parameter. 
For training, $\mbox{\msync}_i$ is used to compute the probability of the next word, using a sampled softmax~\cite{Bengio:03}.
For downstream tasks, we use a learned linear weighting of all layers in the encoders to obtain a task-specific \msync, following \citet{Peters:18}.

\paragraph{Staged parameter updates}
\label{sec:strategy}
Jointly training both the sequential encoder $e_{\mathit{seq}}$, and the syntactic encoder $e_{\mathit{syn}}$ can be expensive, due to the large number of parameters involved.
To reduce cost, we initialize our sequential \cwrs $\seq{h}$, using \textit{pretrained} embeddings from \elmot. 
Once initialized as such, the encoder is fine-tuned to the data likelihood objective~(\S\ref{sec:training}).
% \sect{sec:training} \nascomment{refer to an equation here, not a section}. 
This results in a staged parameter update, which reduces training duration by a factor of 10 in our experiments.
We discuss the empirical effect of this approach in \S\ref{sec:ablations}.

\section{Shallow Syntactic Features}
\label{sec:shallow_features}

Our second approach incorporates shallow syntactic information in downstream tasks via token-level chunk label embeddings. 
Task training (and test) data is automatically chunked, and
chunk boundary information is passed into the task model via BIOUL encoding of the labels.
We add randomly initialized chunk label embeddings to task-specific input encoders, which are then fine-tuned for task-specific objectives.
This approach does not require a shallow syntactic encoder  or chunk annotations for pretraining \cwrs, only a chunker.
Hence, this can more directly measure the impact of shallow syntax for a given task.\footnote{In contrast, in \S\ref{sec:pretraining}, the shallow-syntactic encoder itself, as well as predicted chunk quality on the large pretraining corpus could affect downstream performance.}

\section{Experiments}
\label{sec:experiments}

Our experiments evaluate the effect of shallow syntax, via contextualization (\msync, \S\ref{sec:pretraining}) and features (\S\ref{sec:shallow_features}).
We provide comparisons with four baselines---\elmot \cite{Peters:18b}, our reimplementation of the same, as well as two \cwr-free baselines, with and without shallow syntactic features.
Both \elmot and \msync are trained on the 1B word benchmark corpus \citep{Chelba:13}; the latter also employs chunk annotations~(\sect{sec:chunks}).
Experimental settings are detailed in Appendix~\sect{sec:supp-hyperparameters}.

\subsection{Downstream Task Transfer}
\label{sec:downstream}

We employ four tasks to test the impact of shallow syntax.  
The first three, namely, coarse and fine-grained named entity recognition (NER), and constituency parsing, are \emph{span-based}; the fourth is a sentence-level sentiment classification task.
Following \citet{Peters:18}, we do not apply finetuning to task-specific architectures, allowing us to do a controlled comparison with ELMo.
\todo{Not entirely correct...}
Given an identical base architecture across models for each task, we can attribute any difference in performance to the incorporation of shallow syntax or contextualization.
Details of downstream architectures are provided below, and overall dataset statistics for all tasks is shown in the Appendix, Table \ref{tab:downstream_datas_stats}.

%%%%%%%%%
% DOWNSTREAM
\begin{table*}[tbh]
    \small
	\center
	\begin{tabulary}{\columnwidth}{@{} lrrrr @{}}
		\toprule
		
		& \bf\begin{tabular}[x]{@{}r@{}}NER\end{tabular}
		& \bf\begin{tabular}[x]{@{}r@{}}Fine-grained NER\end{tabular}
		& \bf\begin{tabular}[x]{@{}r@{}}Constituency Parsing\end{tabular}
		& \bf\begin{tabular}[x]{@{}r@{}}Sentiment\end{tabular}\\
		\midrule
		
		Baseline (no \cwr)
		& 88.1 $\pm$ 0.27
		& 78.5 $\pm$ 0.19
		& 88.9 $\pm$ 0.05
		& 51.6 $\pm$ 1.63\\
		
		\quad + shallow syn. features
		& 88.6 $\pm$ 0.22
		& 78.9 $\pm$ 0.13
		& 90.8 $\pm$ 0.14
		& 51.1 $\pm$ 1.39\\
		\midrule[0.03em]
		
	   % \begin{tabular}[x]{@{}r@{}}\textbf{ELMo}\\Transformer\end{tabular}	
	   \elmot~\citep{Peters:18b}
		& 91.1 $\pm$ 0.26 
		& ---
		& 93.7 \textcolor{white}{$\pm$ 0.00}
		& ---\\
		
% 		\begin{tabular}[x]{@{}r@{}}\textbf{ELMo}\\our reimpl.\end{tabular}
        \elmot (our reimplementation)
		& 91.5 $\pm$ 0.25
		& 85.7 $\pm$ 0.08 
		& 94.1 $\pm$ 0.06
		& 53.0 $\pm$ 0.72 \\
		
		\quad + shallow syn. features
		& 91.6 $\pm$ 0.40
		& 85.9 $\pm$ 0.28
		& 94.3 $\pm$ 0.03
		& 52.6 $\pm$ 0.54 \\
		
% 		\begin{tabular}[x]{@{}r@{}}\textbf{\msync}\\ours\end{tabular}
        Shallow syn. contextualization (\msync)
		& 91.5 $\pm$ 0.19 
		& 85.9 $\pm$ 0.20 
		& 94.1 $\pm$ 0.07 
 		& 53.0 $\pm$ 1.07 \\
		
		\bottomrule
	\end{tabulary}
	\caption{
	Test-set performance of \elmot~\cite{Peters:18b}, our reimplementation, and \msync, compared to baselines without \cwr. 
	Evaluation metric is $F_1$ for all tasks except sentiment, which reports accuracy. 
	Reported results show the mean and standard deviation across 5 runs for coarse-grained NER and sentiment classification and 3 runs for other tasks.
% 	\nascomment{by which you mean just parsing?}\swabha{also fine-grained ner}
	%Number of seeds selected based on downstream data sizes \nascomment{footnotes do not work in captions; see how it's missing in the pdf?  I don't think the point in this missing footnote is too important, it's also not very clear what it means}
	}
	\label{tab:downstream-results}
\end{table*}
%%%%%%%%%%

%%%%%%%%%%
% PROBES
\begin{table*}[tbh]
\small
	\center
	\begin{tabulary}{\columnwidth}{@{} l   rrrrrrrrrrr @{}}
		\toprule
% 		& \bf \tabincell{2}{\\Avg.}  
		& \bf CCG
		& \bf \begin{tabular}[x]{@{}r@{}}PTB\\POS\\\end{tabular}
		& \bf \begin{tabular}[x]{@{}r@{}}EWT\\POS\\\end{tabular} 
		& \bf Chunk
		& \bf NER  
		& \begin{tabular}[x]{@{}r@{}}\textbf{Sem.}\\\textbf{Tagging}\\\end{tabular}    
		& \begin{tabular}[x]{@{}r@{}}\textbf{Gramm.}\\\textbf{Err.~D}\\\end{tabular}  
		& \begin{tabular}[x]{@{}r@{}}\textbf{Prep.}\\\textbf{Role}\\\end{tabular}  
		& \begin{tabular}[x]{@{}r@{}}\textbf{Prep.}\\\textbf{Func.}\\\end{tabular}  
		& \begin{tabular}[x]{@{}r@{}}\textbf{Event}\\\textbf{Fact.}\\\end{tabular} \\
		
		\midrule
		
		\elmot
% 		& 79.62 
		& 92.68  
		& 97.09  
		& 95.13  
		& 92.18  
		& 81.21  
		& 93.78  
		& 30.80   
		& 72.81    
		& 82.24   
		& 70.88 \\
		
		\msync
% 		& 79.04 
		& 92.03 
		& 96.91  
		& 94.64 % 94.62  
		& 96.89  %91.00  
		& 79.98 %80.66  
		& 93.03 
		& 30.86  
		& 70.83 %71.71    
		& 82.67  % 82.24 
		& 70.39 \\ %70.76 \\

		\bottomrule
	\end{tabulary}
	\caption{Test performance of \elmot \cite{Peters:18b} vs. \msync on several linguistic probes from \citet{Liu:19}. 
	In each case, performance of the best layer from the architecture is reported. 
	Details on the probes can be found in~\sect{sec:appendixprobe}. }
	\label{tab:probing}
\end{table*}
%%%%%%%%%%

\paragraph{NER} We use the English portion of the CoNLL 2003 dataset \citep{Tjong:03}, which provides named entity annotations on newswire data across four different entity types (\texttt{PER, LOC, ORG, MISC}). 
A bidirectional LSTM-CRF architecture \citep{Lample:16} and a BIOUL tagging scheme were used. 

\paragraph{Fine-grained NER} The same architecture and tagging scheme from above is also used to predict fine-grained entity annotations from OntoNotes 5.0 \citep{Weischedel:11}.
There are 18 fine-grained NER labels in the dataset, including regular named entitities as well as entities such as date, time and common numerical entries.

% \begin{table*}[tbh]
% 	\centering
% 	\begin{tabulary}{\columnwidth}{@{}l   r@{}}
% 		\toprule
% 		\bf Label & \bf Description \\
% 		\midrule[0.03em]
% \texttt{PERSON} & People, including fictional \\
% \texttt{NORP} & Nationalities or religious or political groups \\
% \texttt{FACILITY} & Buildings, airports, highways, bridges, etc. \\
% \texttt{ORGANIZATION} & Companies, agencies, institutions, etc. \\
% \texttt{GPE} & (Geo-Political Entity) Countries, cities, states \\
% \texttt{LOCATION} & Non-GPE locations, mountain ranges, bodies of water \\
% \texttt{PRODUCT} & Vehicles, weapons, foods, etc. (Not services) \\
% \texttt{EVENT} & Named hurricanes, battles, wars, sports events, etc. \\
% \texttt{WORK OF ART} & Titles of books, songs, etc. \\
% \texttt{LAW} & Named documents made into laws  \\
% \texttt{LANGUAGE} & Any named language \\
% % The following values are also annotated in a style similar to names:
% \midrule[0.03em]
% \texttt{DATE} & Absolute or relative dates or periods \\
% \texttt{TIME} & Times smaller than a day \\
% \texttt{PERCENT} & Percentage (including ``\%'') \\
% \texttt{MONEY} & Monetary values, including unit \\
% \texttt{QUANTITY} & Measurements, as of weight or distance \\
% \texttt{ORDINAL} & ``first'', ``second'' \\
% \texttt{CARDINAL} & Numerals that do not fall under another type \\
% \bottomrule
% \end{tabulary}
% 	\caption{Fine-grained NER labels from OntoNotes 5.0 \citep{Weischedel:13}. \todo{Also include frequencies.}}
% 	\label{tab:fgn_labels}
% \end{table*}

\paragraph{Phrase-structure parsing} We use the standard Penn Treebank splits, and adopt the span-based model from \citet{Stern:17}. 
Following their approach, we used predicted part-of-speech tags from the Stanford tagger \citep{Toutanova:03} for training and testing.
About 51\% of phrase-syntactic constituents align exactly with the predicted chunks used, with a majority being single-width noun phrases.
Given that the rule-based procedure used to obtain chunks only propagates the phrase type to the head-word and removes all overlapping phrases to the right, this is expected.
We did not employ jack-knifing to obtain predicted chunks on PTB data; as a result there might be differences in the quality of shallow syntax annotations between the train and test portions of the data.

\paragraph{Sentiment analysis} We consider fine-grained (5-class) classification on Stanford Sentiment Treebank \citep{Socher:13}. 
The labels are \texttt{negative}, \texttt{somewhat\_negative}, \texttt{neutral}, \texttt{positive} and \texttt{somewhat\_positive}.
Our model was based on the biattentive classification network \citep{Mccann:17}.
We used all phrase lengths in the dataset for training, but test results are reported only on full sentences, following prior work.

Results are shown in Table~\ref{tab:downstream-results}.  
Consistent with previous findings, \cwrs offer large improvements across all tasks.  Though helpful to span-level task models without \cwrs, shallow syntactic features  offer little to no benefit to ELMo models.  \msync's performance is similar.
This holds even for phrase-structure parsing, where (gold) chunks align with syntactic phrases, indicating that task-relevant signal learned from exposure to shallow syntax is \emph{already} learned by ELMo.
On sentiment classification, chunk features are slightly harmful on average (but variance is high); \msync again performs similarly to \elmot.
Overall, the performance differences across all tasks are small enough to infer that shallow syntax is not particularly helpful when using  \cwrs.

\subsection{Linguistic Probes}
\label{sec:probes}

We further analyze whether awareness of shallow syntax carries over to other linguistic tasks, via probes from \citet{Liu:19}.
Probes are linear models trained on frozen \cwrs to make predictions about linguistic (syntactic and semantic) properties of words and phrases.
Unlike \S\ref{sec:downstream}, there is minimal downstream task architecture, bringing into focus the transferability of \cwrs, as opposed to task-specific adaptation.

\subsubsection{Probing Tasks}
\label{sec:appendixprobe}

The ten different probing tasks we used include CCG supertagging \cite{Hockenmaier:07}, part-of-speech tagging from PTB \cite{Marcus:93} and  EWT (Universal Depedencies \citealp{Silveira:14}), named entity recognition \cite{Tjong:03}, base-phrase chunking \cite{Tjong:00}, grammar error detection \cite{Yannakoudakis:11}, semantic tagging \cite{Bjerva:16}, preposition supersense identification \cite{Schneider:18}, and event factuality detection \cite{Rudinger:18}.
Metrics and references for each are summarized in Table~\ref{tab:moreprobing}. For more details, please see \citet{Liu:19}.

Results in Table~\ref{tab:probing} show ten probes.
%---CCG supertagging, part-of-speech tagging on PTB and  Universal Dependencies, chunking, NER, grammar error detection, semantic tagging, preposition supersense identification, and event factuality detection
Again, we see the performance of baseline \elmot~and \msync are similar, with \msync doing slightly worse on 7 out of 9 tasks. 
As we would expect, on the probe for predicting chunk tags, \msync achieves 96.9 $F_1$ vs.~92.2 $F_1$ for \elmot, indicating that \msync is indeed encoding shallow syntax. 
Overall, the results further confirm that explicit shallow syntax does not offer any benefits over \elmot.

\subsection{Effect of Training Scheme}
\label{sec:ablations}
We test whether our staged parameter training~(\sect{sec:strategy}) is a viable alternative to an end-to-end training of both $e_{\mathit{syn}}$ and $e_{\mathit{seq}}$.
We make a further distinction between fine-tuning $e_{\mathit{seq}}$ vs.~not updating it at all after initialization (frozen).
%%%%%%%%%%
% ABLATIONS
\begin{table}[tbh]
\small
	\center
	\begin{tabulary}{\columnwidth}{@{}ll   r@{}}
		\toprule

		& \textbf{Model}
		& \begin{tabular}[x]{@{}r@{}}\textbf{Fine-grained}\\\textbf{NER $F_1$}\end{tabular}\\
		
		\midrule
		
		\multirow{2}{*}[-3pt]{end-to-end}
		& ELMo 
		& 86.90 $\pm$ 0.11 \\
		
		\cmidrule{2-3}
		&\msync end-to-end 
		& 86.89 $\pm$ 0.04 \\
		
		\midrule
		\multirow{2}{*}[-3pt]{staged}
		& \msync frozen
		& 87.36 $\pm$ 0.02 \\
		
		&\msync fine-tuned
		& 87.44 $\pm$ 0.07 \\

		\bottomrule
	\end{tabulary}
	\caption{Validation $F_1$ for fine-grained NER across syntactic pretraining schemes, with mean and standard deviations across 3 runs.}
	\label{tab:pretraining}
\end{table}
%%%%%%%%%%
Downstream validation-set $F_1$ on fine-grained NER, reported in Table \ref{tab:pretraining}, shows that the end-to-end strategy lags behind the others, perhaps indicating the need to train longer than 10 epochs. 
However, a single epoch on the 1B-word benchmark takes 36 hours on 2 Tesla V100s, making this prohibitive.
Interestingly, the frozen strategy, which takes the least amount of time to converge (24 hours on 1 Tesla V100), also performs almost as well as fine-tuning.

\section{Conclusion}
We find that exposing \cwr-based models to shallow syntax, either through new \cwr learning architectures or explicit pipelined features, has little effect on their performance, across several tasks.
Linguistic probing also shows that \cwrs aware of such structures do not improve task transferability.
Our architecture and methods are general enough to be adapted for  richer inductive biases, such as those given by full syntactic trees (RNNGs; \citealp{Dyer:16}), or to different pretraining objectives, such as masked language modeling (BERT; \citealp{Devlin:18}); we leave this pursuit to future work.

\bibliography{references}

\begin{thebibliography}{35}
\expandafter\ifx\csname natexlab\endcsname\relax\def\natexlab#1{#1}\fi

\bibitem[{Abney(1991)}]{Abney:91}
Steven~P Abney. 1991.
\newblock \href
  {https://link.springer.com/chapter/10.1007/978-94-011-3474-3_10} {Parsing by
  chunks}.
\newblock In \emph{Principle-based parsing}, pages 257--278. Springer.

\bibitem[{Bengio et~al.(2003)Bengio, Ducharme, Vincent, and Janvin}]{Bengio:03}
Yoshua Bengio, R{\'e}jean Ducharme, Pascal Vincent, and Christian Janvin. 2003.
\newblock \href {http://dl.acm.org/citation.cfm?id=944919.944966} {A neural
  probabilistic language model}.
\newblock \emph{Journal of Machine Learning Research}, pages 1137--1155.

\bibitem[{Bjerva et~al.(2016)Bjerva, Plank, and Bos}]{Bjerva:16}
Johannes Bjerva, Barbara Plank, and Johan Bos. 2016.
\newblock \href {https://www.aclweb.org/anthology/C16-1333} {Semantic tagging
  with deep residual networks}.
\newblock In \emph{Proc. of COLING}.

\bibitem[{Blevins et~al.(2018)Blevins, Levy, and Zettlemoyer}]{Blevins:18}
Terra Blevins, Omer Levy, and Luke Zettlemoyer. 2018.
\newblock \href {https://aclweb.org/anthology/P18-2003} {Deep rnns encode soft
  hierarchical syntax}.
\newblock In \emph{Proc. of ACL}.

\bibitem[{Chelba et~al.(2013)Chelba, Mikolov, Schuster, Ge, Brants, Koehn, and
  Robinson}]{Chelba:13}
Ciprian Chelba, Tomas Mikolov, Mike Schuster, Qi~Ge, Thorsten Brants, Phillipp
  Koehn, and Tony Robinson. 2013.
\newblock \href {https://arxiv.org/abs/1312.3005} {One billion word benchmark
  for measuring progress in statistical language modeling}.
\newblock ArXiv:1312.3005.

\bibitem[{Devlin et~al.(2018)Devlin, Chang, Lee, and Toutanova}]{Devlin:18}
Jacob Devlin, Ming{-}Wei Chang, Kenton Lee, and Kristina Toutanova. 2018.
\newblock \href {http://arxiv.org/abs/1810.04805} {{BERT:} pre-training of deep
  bidirectional transformers for language understanding}.
\newblock ArXiv:1810.04805.

\bibitem[{Dyer et~al.(2016)Dyer, Kuncoro, Ballesteros, and Smith}]{Dyer:16}
Chris Dyer, Adhiguna Kuncoro, Miguel Ballesteros, and Noah~A Smith. 2016.
\newblock \href {https://www.aclweb.org/anthology/N16-1024} {Recurrent neural
  network grammars}.
\newblock In \emph{Proc. of NAACL-HLT}.

\bibitem[{Gardner et~al.(2017)Gardner, Grus, Neumann, Tafjord, Dasigi, Liu,
  Peters, Schmitz, and Zettlemoyer}]{Gardner:17}
Matt Gardner, Joel Grus, Mark Neumann, Oyvind Tafjord, Pradeep Dasigi,
  Nelson~F. Liu, Matthew Peters, Michael Schmitz, and Luke~S. Zettlemoyer.
  2017.
\newblock \href {https://arxiv.org/abs/1803.07640} {{AllenNLP}: A deep semantic
  natural language processing platform}.
\newblock ArXiv:1803.07640.

\bibitem[{Hockenmaier and Steedman(2007)}]{Hockenmaier:07}
Julia Hockenmaier and Mark Steedman. 2007.
\newblock \href {https://www.aclweb.org/anthology/J07-3004} {{CCGbank}: A
  corpus of ccg derivations and dependency structures extracted from the {Penn
  Treebank}}.
\newblock \emph{Computational Linguistics}, 33(3).

\bibitem[{Howard and Ruder(2018)}]{Howard:18}
Jeremy Howard and Sebastian Ruder. 2018.
\newblock \href {http://www.aclweb.org/anthology/P18-1031} {Universal language
  model fine-tuning for text classification}.
\newblock In \emph{Proc. of ACL}.

\bibitem[{Jurafsky and Martin(2000)}]{Jurafsky:00}
Daniel Jurafsky and James~H. Martin. 2000.
\newblock \href {https://dl.acm.org/citation.cfm?id=555733} {\emph{Speech and
  Language Processing: An Introduction to Natural Language Processing,
  Computational Linguistics, and Speech Recognition}}, 1st edition.
\newblock Prentice Hall PTR, Upper Saddle River, NJ, USA.

\bibitem[{Kim et~al.(2016)Kim, Jernite, Sontag, and
  Rush}]{Kim:2016:CNL:3016100.3016285}
Yoon Kim, Yacine Jernite, David Sontag, and Alexander~M. Rush. 2016.
\newblock \href {http://dl.acm.org/citation.cfm?id=3016100.3016285}
  {Character-aware neural language models}.
\newblock In \emph{Proceedings of the Thirtieth AAAI Conference on Artificial
  Intelligence}, AAAI'16, pages 2741--2749. AAAI Press.

\bibitem[{Kuncoro et~al.(2018)Kuncoro, Dyer, Hale, Yogatama, Clark, and
  Blunsom}]{Kuncoro:18}
Adhiguna Kuncoro, Chris Dyer, John Hale, Dani Yogatama, Stephen Clark, and Phil
  Blunsom. 2018.
\newblock \href {https://aclweb.org/anthology/P18-1132} {Lstms can learn
  syntax-sensitive dependencies well, but modeling structure makes them
  better}.
\newblock In \emph{Proc. of ACL}.

\bibitem[{Lample et~al.(2016)Lample, Ballesteros, Subramanian, Kawakami, and
  Dyer}]{Lample:16}
Guillaume Lample, Miguel Ballesteros, Sandeep Subramanian, Kazuya Kawakami, and
  Chris Dyer. 2016.
\newblock \href {https://www.aclweb.org/anthology/N16-1030} {Neural
  architectures for named entity recognition}.
\newblock In \emph{Proc. of NAACL-HLT}.

\bibitem[{Liu et~al.(2019)Liu, Gardner, Belinkov, Peters, and Smith}]{Liu:19}
Nelson~F. Liu, Matt Gardner, Yonatan Belinkov, Matthew Peters, and Noah~A.
  Smith. 2019.
\newblock \href {https://arxiv.org/abs/1903.08855} {Linguistic knowledge and
  transferability of contextual representations}.
\newblock In \emph{Proc. of NAACL-HLT}.

\bibitem[{Marcus et~al.(1993)Marcus, Marcinkiewicz, and Santorini}]{Marcus:93}
Mitchell~P Marcus, Mary~Ann Marcinkiewicz, and Beatrice Santorini. 1993.
\newblock \href {http://aclweb.org/anthology/J93-2004} {Building a large
  annotated corpus of {English}: The {Penn} treebank}.
\newblock \emph{Computational Linguistics}, 19(2):313--330.

\bibitem[{McCann et~al.(2017)McCann, Bradbury, Xiong, and Socher}]{Mccann:17}
Bryan McCann, James Bradbury, Caiming Xiong, and Richard Socher. 2017.
\newblock \href
  {http://papers.nips.cc/paper/7209-learned-in-translation-contextualized-word-vectors}
  {Learned in translation: Contextualized word vectors}.
\newblock In \emph{Proc. of NeurIPS}, pages 6294--6305.

\bibitem[{Peters et~al.(2017)Peters, Ammar, Bhagavatula, and Power}]{Peters:17}
Matthew Peters, Waleed Ammar, Chandra Bhagavatula, and Russell Power. 2017.
\newblock \href {https://aclweb.org/anthology/papers/P/P17/P17-1161}
  {Semi-supervised sequence tagging with bidirectional language models}.
\newblock In \emph{Proc. of ACL}.

\bibitem[{Peters et~al.(2018{\natexlab{a}})Peters, Neumann, Iyyer, Gardner,
  Clark, Lee, and Zettlemoyer}]{Peters:18}
Matthew Peters, Mark Neumann, Mohit Iyyer, Matt Gardner, Christopher Clark,
  Kenton Lee, and Luke Zettlemoyer. 2018{\natexlab{a}}.
\newblock \href {http://www.aclweb.org/anthology/N18-1202} {Deep contextualized
  word representations}.
\newblock In \emph{Proc. of NAACL-HLT}.

\bibitem[{Peters et~al.(2018{\natexlab{b}})Peters, Neumann, Zettlemoyer, and
  Yih}]{Peters:18b}
Matthew Peters, Mark Neumann, Luke Zettlemoyer, and Wen-tau Yih.
  2018{\natexlab{b}}.
\newblock \href {http://www.aclweb.org/anthology/D18-1179} {Dissecting
  contextual word embeddings: Architecture and representation}.
\newblock In \emph{Proc. of EMNLP}, pages 1499--1509.

\bibitem[{Radford et~al.(2018)Radford, Narasimhan, Salimans, and
  Sutskever}]{Radford:18}
Alec Radford, Karthik Narasimhan, Tim Salimans, and Ilya Sutskever. 2018.
\newblock \href
  {https://www.cs.ubc.ca/~amuham01/LING530/papers/radford2018improving.pdf}
  {Improving language understanding by generative pre-training}.

\bibitem[{Rudinger et~al.(2018)Rudinger, White, and Van~Durme}]{Rudinger:18}
Rachel Rudinger, Aaron~Steven White, and Benjamin Van~Durme. 2018.
\newblock \href {https://www.aclweb.org/anthology/N18-1067} {Neural models of
  factuality}.
\newblock In \emph{Proc. of ACL}.

\bibitem[{Schneider et~al.(2018)Schneider, Hwang, Srikumar, Prange, Blodgett,
  Moeller, Stern, Bitan, and Abend}]{Schneider:18}
Nathan Schneider, Jena~D Hwang, Vivek Srikumar, Jakob Prange, Austin Blodgett,
  Sarah~R Moeller, Aviram Stern, Adi Bitan, and Omri Abend. 2018.
\newblock \href {https://www.aclweb.org/anthology/P18-1018} {Comprehensive
  supersense disambiguation of {English} prepositions and possessives}.

\bibitem[{Silveira et~al.(2014)Silveira, Dozat, De~Marneffe, Bowman, Connor,
  Bauer, and Manning}]{Silveira:14}
Natalia Silveira, Timothy Dozat, Marie-Catherine De~Marneffe, Samuel~R Bowman,
  Miriam Connor, John Bauer, and Christopher~D Manning. 2014.
\newblock \href
  {http://www.lrec-conf.org/proceedings/lrec2014/pdf/1089_Paper.pdf} {A gold
  standard dependency corpus for {English}.}
\newblock In \emph{Proc. of LREC}, pages 2897--2904.

\bibitem[{Socher et~al.(2013)Socher, Perelygin, Wu, Chuang, Manning, Ng, and
  Potts}]{Socher:13}
Richard Socher, Alex Perelygin, Jean Wu, Jason Chuang, Christopher~D Manning,
  Andrew Ng, and Christopher Potts. 2013.
\newblock \href {https://aclweb.org/anthology/papers/D/D13/D13-1170} {Recursive
  deep models for semantic compositionality over a sentiment treebank}.
\newblock In \emph{Proc. of EMNLP}.

\bibitem[{Stern et~al.(2017)Stern, Andreas, and Klein}]{Stern:17}
Mitchell Stern, Jacob Andreas, and Dan Klein. 2017.
\newblock \href {https://www.aclweb.org/anthology/papers/P/P17/P17-1076/} {A
  minimal span-based neural constituency parser}.
\newblock In \emph{Proc. of ACL}.

\bibitem[{Strubell et~al.(2018)Strubell, Verga, Andor, Weiss, and
  McCallum}]{Strubell:18}
Emma Strubell, Patrick Verga, Daniel Andor, David Weiss, and Andrew McCallum.
  2018.
\newblock \href {https://aclweb.org/anthology/D18-1548}
  {Linguistically-informed self-attention for semantic role labeling}.
\newblock In \emph{Proc. of EMNLP}.

\bibitem[{Tenney et~al.(2019)Tenney, Xia, Chen, Wang, Poliak, McCoy, Kim,
  Durme, Bowman, Das, and Pavlick}]{Tenney:18}
Ian Tenney, Patrick Xia, Berlin Chen, Alex Wang, Adam Poliak, R~Thomas McCoy,
  Najoung Kim, Benjamin~Van Durme, Sam Bowman, Dipanjan Das, and Ellie Pavlick.
  2019.
\newblock \href {https://openreview.net/forum?id=SJzSgnRcKX} {What do you learn
  from context? {Probing} for sentence structure in contextualized word
  representations}.
\newblock In \emph{Proc. of ICLR}.

\bibitem[{Tjong Kim~Sang and Buchholz(2000)}]{Tjong:00}
Erik~F. Tjong Kim~Sang and Sabine Buchholz. 2000.
\newblock \href {http://www.aclweb.org/anthology/W00-0726} {Introduction to the
  {CoNLL}-2000 shared task: Chunking}.
\newblock In \emph{Proc. of CoNLL}.

\bibitem[{Tjong Kim~Sang and De~Meulder(2003)}]{Tjong:03}
Erik~F Tjong Kim~Sang and Fien De~Meulder. 2003.
\newblock \href {https://aclweb.org/anthology/W03-0419} {Introduction to the
  {CoNLL}-2003 shared task: Language-independent named entity recognition}.
\newblock In \emph{Proc. of NAACL}. Association for Computational Linguistics.

\bibitem[{Toutanova et~al.(2003)Toutanova, Klein, Manning, and
  Singer}]{Toutanova:03}
Kristina Toutanova, Dan Klein, Christopher~D. Manning, and Yoram Singer. 2003.
\newblock \href {https://doi.org/10.3115/1073445.1073478} {Feature-rich
  part-of-speech tagging with a cyclic dependency network}.
\newblock In \emph{Proc. of NAACL}.

\bibitem[{Vaswani et~al.(2017)Vaswani, Shazeer, Parmar, Uszkoreit, Jones,
  Gomez, Kaiser, and Polosukhin}]{Vaswani:17}
Ashish Vaswani, Noam Shazeer, Niki Parmar, Jakob Uszkoreit, Llion Jones,
  Aidan~N Gomez, {\L}ukasz Kaiser, and Illia Polosukhin. 2017.
\newblock \href
  {https://papers.nips.cc/paper/7181-attention-is-all-you-need.pdf} {Attention
  is all you need}.
\newblock In \emph{Proc. of NeurIPS}, pages 5998--6008.

\bibitem[{Weischedel et~al.(2013)Weischedel, Palmer, Marcus, Hovy, Pradhan,
  Ramshaw, Xue, Taylor, Kaufman, Franchini et~al.}]{Weischedel:13}
Ralph Weischedel, Martha Palmer, Mitchell Marcus, Eduard Hovy, Sameer Pradhan,
  Lance Ramshaw, Nianwen Xue, Ann Taylor, Jeff Kaufman, Michelle Franchini,
  et~al. 2013.
\newblock \href {https://catalog.ldc.upenn.edu/LDC2013T19} {{OntoNotes} release
  5.0 ldc2013t19}.
\newblock \emph{Linguistic Data Consortium, Philadelphia, PA}.

\bibitem[{Weischedel et~al.(2011)Weischedel, Pradhan, Ramshaw, Palmer, Xue,
  Marcus, Taylor, Greenberg, Hovy, Belvin et~al.}]{Weischedel:11}
Ralph Weischedel, Sameer Pradhan, Lance Ramshaw, Martha Palmer, Nianwen Xue,
  Mitchell Marcus, Ann Taylor, Craig Greenberg, Eduard Hovy, Robert Belvin,
  et~al. 2011.
\newblock \href
  {https://catalog.ldc.upenn.edu/docs/LDC2011T03/OntoNotes-Release-4.0.pdf}
  {{OntoNotes Release 4.0}}.
\newblock \emph{LDC2011T03, Philadelphia, Penn.: Linguistic Data Consortium}.

\bibitem[{Yannakoudakis et~al.(2011)Yannakoudakis, Briscoe, and
  Medlock}]{Yannakoudakis:11}
Helen Yannakoudakis, Ted Briscoe, and Ben Medlock. 2011.
\newblock \href {https://www.aclweb.org/anthology/P11-1019} {A new dataset and
  method for automatically grading esol texts}.
\newblock In \emph{Proc. of ACL}.

\end{thebibliography}
\bibliographystyle{acl_natbib}

\appendix
\section{Supplemental Material}
\label{sec:supplemental}

\subsection{Hyperparameters}
\label{sec:supp-hyperparameters}

\paragraph{\elmot} Our baseline pretraining model was a reimplementation of that given in \citet{Peters:18b}. 
Hyperparameters were generally identical, but we trained on only 2 GPUs with (up to) 4,000 tokens per batch. 
This difference in batch size meant we used 6,000 warm up steps with the learning rate schedule of \citet{Vaswani:17}. 

\paragraph{\msync} The function $f_{seq}$ is identical to the 6-layer biLM used in \elmot. $f_{syn}$, on the other hand, uses only 2 layers. The learned embeddings for the chunk labels have 128 dimensions and are concatenated with the two boundary $\seq{h}$ of dimension 512. Thus $f_{proj}$ maps $1024 + 128$ dimensions to 512. Further, we did not perform weight averaging over several checkpoints.

\paragraph{Shallow Syntax} The size of the shallow syntactic feature embedding was 50 across all experiments, initialized uniform randomly.

All model implementations are based on the \ai~library \cite{Gardner:17}.

\begin{table*}[tbh]
	\centering
	\begin{tabulary}{\columnwidth}{@{}l   rrr@{}}
		\toprule
		\bf Task &  \bf Train  & \bf Heldout & \bf Test\\
		\midrule[0.03em]
		CoNLL 2003 NER \citep{Tjong:03}     & 23,499 &  5,942 & 5,648 \\
		OntoNotes NER \citep{Weischedel:13} & 81,828 & 11,066 & 11,257 \\
		Penn TreeBank \citep{Marcus:93}     & 39,832 &  1,700 & 2,416  \\
		Stanford Sentiment Treebank \citep{Socher:13} & 8,544 & 1,101 & 2,210 \\
		\bottomrule
\end{tabulary}
	\caption{Downstream dataset statistics describing the number of train, heldout and test set instances for each task. \todo{Verify!}}
	\label{tab:downstream_datas_stats}
\end{table*}

\begin{table*}[tbh]
	\center
	\begin{tabulary}{\columnwidth}{@{}lrr@{}}

		\toprule
		Task
		& Dataset 
	    & Metric \\
		\midrule
		
		CCG Supertagging
		& CCGBank \cite{Hockenmaier:07}
		& Accuracy\\
		
		PTB part-of-speech tagging
		& PennTreeBank \cite{Marcus:93}
		& Accuracy\\
		
		EWT part-of-speech tagging
		& Universal Dependencies \cite{Silveira:14}
		& Accuracy\\
		
		Chunking
		& CoNLL 2000 \cite{Tjong:00}
		& $F_1$\\
		
		Named Entity Recognition
		& CoNLL 2003 \cite{Tjong:03}
		& $F_1$
		\\
		
		Semantic Tagging
		& \cite{Bjerva:16}
		& Accuracy
		\\
		
		Grammar Error Detection
		& First Certificate in English \cite{Yannakoudakis:11}
		& $F_1$
		\\
		
		Preposition Supersense Role
		& STREUSLE 4.0 \cite{Schneider:18}
		& Accuracy
		\\
		
		Preposition Supersense Function
		& STREUSLE 4.0 \cite{Schneider:18}
		& Accuracy
		\\
		
		Event Factuality Detection
		& UDS It Happened v2 \cite{Rudinger:18}
		& Pearson R
		\\
		
		\bottomrule
	\end{tabulary}
	\caption{Dataset and metrics for each probing task from \citet{Liu:19}, corresponding to Table~\ref{tab:probing}.}
	\label{tab:moreprobing}
\end{table*}

\end{document}